\documentclass[letterpaper, 10 pt, conference]{ieeeconf}  
\pdfoutput=1

\IEEEoverridecommandlockouts 
\usepackage{hyperref}
\usepackage{float}
\usepackage[utf8]{inputenc}
\usepackage{graphicx}
\usepackage{subcaption}
\usepackage{amsmath}
\usepackage{tikz}
\usepackage{pgfplots}
\usepgfplotslibrary{dateplot}
\usepackage{adjustbox}
\usepackage{tabularx,ragged2e,booktabs,caption}
\newcolumntype{L}{>{\RaggedRight\arraybackslash}X}

\usepackage[backend=biber,style=numeric,sorting=none]{biblatex}
\title{\LARGE \bf
OpenUAV Cloud Testbed: a Collaborative Design Studio for Field Robotics}

\author{Harish Anand, Stephen A. Rees, Zhiang Chen, Ashwin Jose, Sarah Bearman,  \\%
Prasad Antervedi, Jnaneshwar Das
{\small
\thanks{*This work was supported by NSF award CNS-1521617}
\thanks{Authors are with the School of Earth and Space Exploration, Tempe, Arizona {hanand4,zchen256, aporuthu,sbearman,lanterve,jdas5@asu.edu} and Stephen is a staff at Vanderbilt University, Nashville, Tennessee {stephen.a.rees@vanderbilt.edu}}%
}}
\begin{document}


\maketitle
\thispagestyle{empty}
\pagestyle{empty}

\begin{abstract}
%


Simulations play a crucial role in robotics research and education. This paper presents the OpenUAV testbed, an open-source, easy-to-use, web-based, and reproducible software system that enables students and researchers to run robotic simulations on the cloud. We have built upon our previous work and have addressed some of the educational and research challenges associated with the prior work. The critical contributions of the paper to the robotics and automation community are threefold:
First, OpenUAV saves students and researchers from tedious and complicated software setups by providing web-browser-based Linux desktop sessions with standard robotics software like Gazebo, ROS, and flight autonomy stack.
Second, a method for saving an individual's research work with its dependencies for the work's future reproducibility.
Third, the platform provides a mechanism to support photorealistic robotics simulations by combining Unity game engine-based camera rendering and Gazebo physics. The paper addresses a research need for photorealistic simulations and describes a methodology for creating a photorealistic aquatic simulation. We also present the various academic and research use-cases of this platform to improve robotics education and research, especially during times like the COVID-19 pandemic, when virtual collaboration is necessary. 

\href{https://github.com/Open-UAV/openuav-turbovnc}{GitHub: https://github.com/Open-UAV/openuav-turbovnc}

\href{https://openuav.us}{Webpage: https://openuav.us}
\end{abstract}

\section{Introduction}


The increasing shift in computing power to cloud-based services is already evident in the gaming industry (Steam and Google Stadia), Over-the-Top (OTT) content services (Netflix, Hulu), video conferencing software (Skype), and workplace communication platforms like Slack \cite{steam, stadia, adhikari2014measurement, baset2004analysis, johnson2018slack}. This push is especially relevant in the current challenging times, such as the COVID-19 pandemic, where remote work and virtual collaborations are necessary for robotics research and education.  


Robotics is one of those challenging fields where there is always an interplay between hardware and software development. In addition to getting all the hardware components to function together correctly, the robotics engineers also face challenging software setup and various task-specific configurations. These software setup challenges increase tremendously when many students and researchers engage in robotics software development using their computers, which typically have different operating systems and software versions. Hence we have presented a cloud-based solution to provide researchers and students with an easy-to-use, reproducible, and web browser-based robotics simulation that can easily offer a collaborative software development experience.

OpenUAV is primarily targeted towards students and researchers familiar with the robotics environment but not having access to significant computational resources like GPUs and CPUs. Our open-source platform is also beneficial for robotics coursework, where all the students have a standard workspace irrespective of their machine operating systems. However, we acknowledge that this approach might hinder a beginners' learning curve from installing and configuring their robotics workflow.

There is a growing demand for Unmanned Aircraft Systems (UAS) based imaging technology to provide high-resolution spatial imaging for science applications \cite{lottes2017uav,ammour2017deep}. For example, UAS-based imaging technology is used in geology to generate high-resolution orthomosaics of fault scarps to study the geomorphological surface processes \cite{chen2019geomorphological}. These imaging technologies can generate high-resolution object models that can be used to create photorealistic simulation environments such as aquatic environments with coral reef models \cite{burns2015integrating}. OpenUAV uses Unity game engine, an easy-to-use, photorealistic rendering software to render high-resolution object models and their environments for robotic simulations \cite{unity}.

\begin{figure*} 

\centering
\vspace{6pt}
\includegraphics[width=\textwidth]{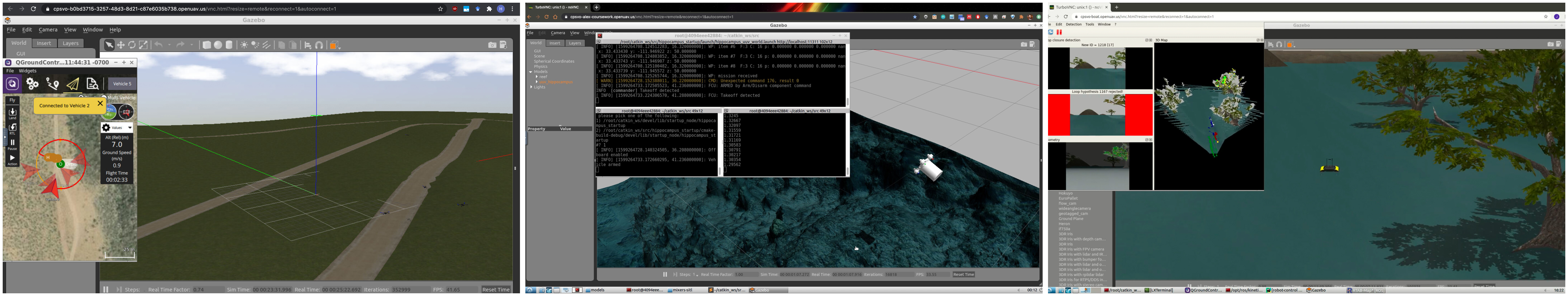}
\caption{A multi-UAS simulation controlled using QGroundControl is shown on the left panel. In the center panel, an underwater coral reef world and an underwater vehicle is present. The right panel shows with RTAB-Map running on a Heron boat with a flight autonomy stack \cite{labbe2019rtab, jdboat}.}
\label{fig:term}
\end{figure*}

We also present the NSF CPS Challenge 2020, a virtual competition conducted in the OpenUAV platform that had teams from different parts of the world compete on the challenge of autonomously deploying and retrieving a soil probe using a UAS and landing the UAS on a moving rover. We found that our web browser-based simulation platform provided an easy alternative to the usual outdoor competition. It provided every team with access to GPU resources and ample computing power. Three examples of OpenUAV simulations accessed through a web browser are shown in Fig.~\ref{fig:term}.



Our platform also allows users to save their research experiments, with all the necessary configuration as a compressed file (docker image), which provides a means to reproduce research works.

\section{Related Work}
Various robotic simulation packages exist for vehicle simulations, and we take a look at some of those works that also achieve photorealism and cloud simulation capabilities. The following are some examples. 

\subsection{OpenUAV1}

The purpose of the earlier version of OpenUAV (referred to as OpenUAV1) was to reduce the barrier to entry in UAS research and education \cite{schmittle2018openuav_1}. The new version address some of the educational and research challenges we faced in achieving the above purpose. 
\begin{itemize}
    \item The beginner-level students typically require a hands-on live session to help them with their configuration issues and questions. We found that a remote desktop configured with flight autonomy software and a widely used Integrated Development Environment (IDE) enables them to quickly start robotics programming. The Gzweb based approach followed in OpenUAV1 hinder beginners and is much more suitable for advanced users \cite{gzweb}.
    
    \item Photorealistic environments that are very similar to outdoor light characteristics are necessary for developing robust vision-based algorithms. Our architecture address this research need by using the Unity game engine's rendering capability and its comparative ease of use with other game engines \cite{dickson2017experience}.

    \item A higher-level mission planning and flight control software is widely used in UAS field experiments. This software is required for researchers working with UAS development. We incorporated QGroundControl, a mission planning software used in PX4 vehicles, to satisfy this need, thus resembling an outdoor field experiment scenario \cite{meier2015px4, zurich2013qgroundcontrol}.
    
    \item We improved our server hardware to an 80 and 64 CPU threaded machine with two NVIDIA TITAN RTX and two NVIDIA GeForce RTX 2080 Ti. The higher CPU threads help users do swarm simulations consisting of 10 or more UAS. Each UAS typically creates three processes, i.e., PX4, ROS, and a controller code \cite{meier2015px4, quigley2009ros}.

\end{itemize}

\subsection{AirSim}
AirSim is an open-source, cross-platform simulator built on Unreal Engine 4 (UE4) that offers similar photorealistic simulations for drones and cars \cite{shah2018airsim}. In OpenUAV's photorealistic simulations, Unity game objects are created using the robot models described in Unified Robot Description Format (URDF) or Software Definition Format (SDF) format and hence support wide range of robots. 

\subsection{FlightGoggles}
FlightGoggles is a simulator that renders a virtual-reality environment for aerial vehicles \cite{guerra2019flightgoggles}. In FlightGoggles, the sensor measurements are artificially rendered in real-time, while the vehicle vibrations and unsteady aerodynamics are from the vehicle's natural interactions. This is an exceptional advantage of the FlightGoggles framework, where a combination of vehicle-in-the-loop physics with the Unity-based rendering of the camera's environment occurs. Photorealistic OpenUAV simulations use Unity to render the cameras with the physics from the Gazebo simulation, inspired by the FlightGoggles approach. However, our work is geared towards achieving a convenient cloud-based UAS simulation for students and researchers supporting photorealism.

\subsection{The Construct Sim}
The Construct Sim is an online robotics teaching platform that provides web-based access to robotics simulators like Gazebo with the Robot Operating System (ROS) support \cite{tellez2017thousand}. They utilize Gzweb, a WebGL client for Gazebo, for developing simulations. Some of the disadvantages we identified while using Gzweb in OpenUAV1 include the lack of an IDE for programming and development, reduced capabilities compared to a Gazebo user interface, and lack of support for file transfer. By replacing gzweb with a remote desktop that can run Gazebo, we make it closer to how they would operate on their local machine. Our goals are similar to the Construct Sim, where we want to provide robotics students with an online, easily accessible simulation platform without getting bogged down by software configuration issues.

\subsection{Robotarium}
The Robotarium is a multi-robot research testbed where users can have remote access to a state-of-the-art multi-robot facility \cite{pickem2017robotarium}. Their approach of sharing the physical robots comes with its challenges like automatic charging and tracking. A user code submitted to their system undergoes a simulation-based verification before running on the physical robots. OpenUAV1 followed a similar approach where a user submits their code, and its results are visualized in Gzweb. However, this approach is replaced in the current version, as students would prefer to do their programming within their remote session.  


\section{Design Goals}
As a remotely accessible, open-source simulation platform, OpenUAV’s main purpose has been to lower the barrier to entry in research and development of autonomous vehicles. To this end, we aim to satisfy the following requirements:

\begin{itemize}  
	\item \textbf{Easy-to-use:} Provide students and researchers with a remote desktop consisting of necessary robotics and programming software to simulate autonomous vehicles.
	
	\item \textbf{Reproducible:} Develop the capability to store student and research work as compressed files to be used in the future to recreate the same experiment.
	
	\item \textbf{Web-based:} Provide remote desktop sessions accessible through web browsers to limit students' local machine requirements to a bare minimum.
	
	\item \textbf{Photorealistic:} Provide photorealistic outdoor environments for field robotic simulations.
\end{itemize}


We found further requirements like daily backups of student and research work and support IDEs like Sublime Text and PyCharm for ease of use during student interactions \cite{sublime, pycharm}. Another concern that students raised was that the default Lubuntu web browser (Firefox) inside the container.

\section{System Design}
\begin{figure} 
\centering
\vspace{6pt}
\includegraphics[width=0.48\textwidth]{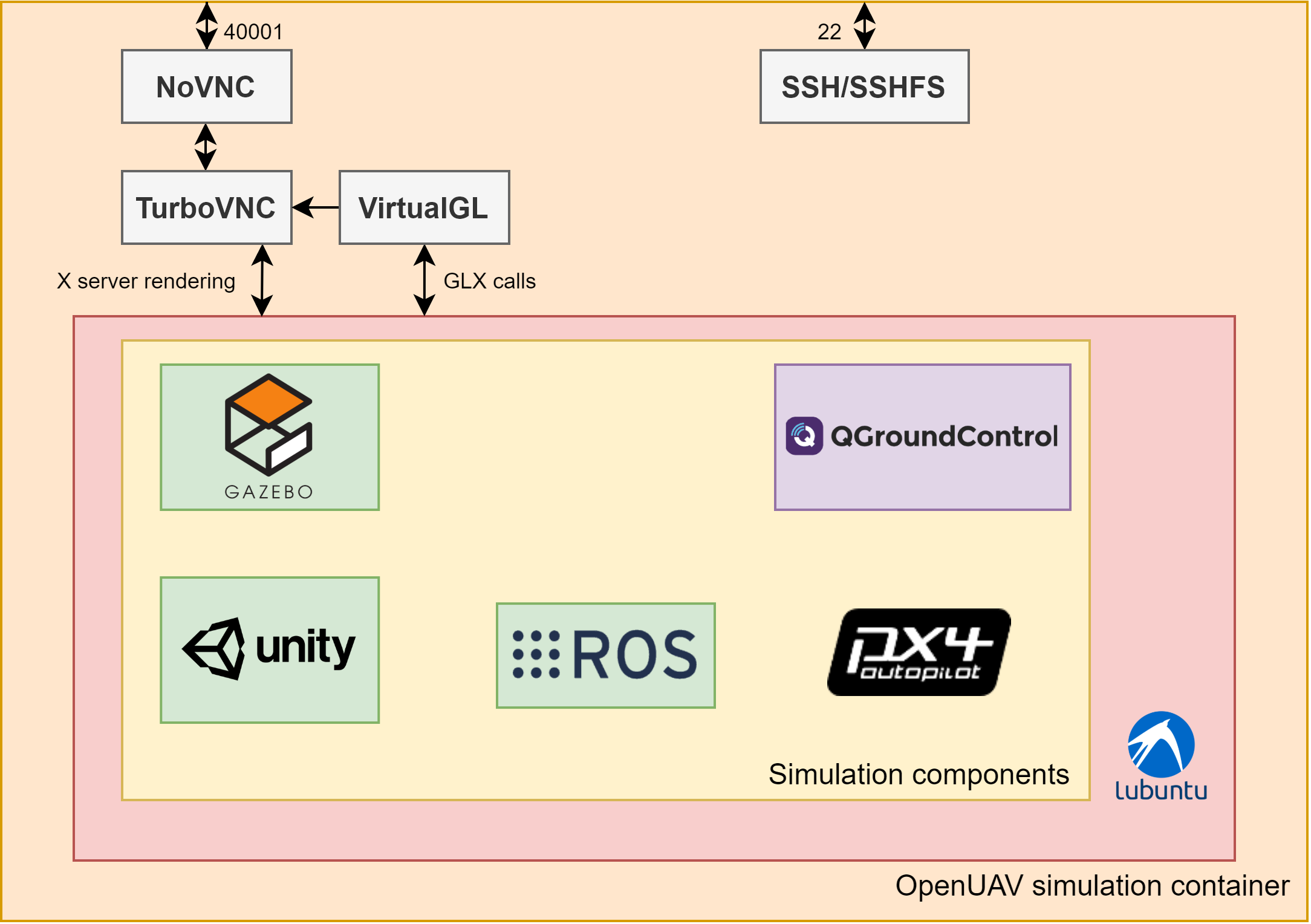}
\caption{Components of an OpenUAV container: The simulation components include Gazebo, Unity, ROS, PX4, and QGroundControl. The user-interactive components consist of NoVNC, TurboVNC, and SSH/SSHFS. NoVNC and SSHFS are accessible on the OpenUAV's machine using the container's IP address with ports 40001 and 22.}
\label{fig:openuav}
\end{figure}

We classify the software components into three categories: simulation components, virtualization components, and interactive components. This updated system design was inspired by the works of Will Kessler and the Udacity team \cite{willkessler}. An overview of the latest OpenUAV simulation container is shown in Fig.~\ref{fig:openuav}.

\subsection{Simulation Components} Simulation has always been a necessity to solve real-world problems safely and efficiently. The software packages that enable simulation in OpenUAV are the following:

\subsubsection{Gazebo and ROS} Gazebo is an open-source robotics software used to design robots and environments and perform realistic rigid body dynamic simulations \cite{koenig2004design}. The Robot Operating System (ROS) is a robotics message passing framework designed to simplify programming for robots \cite{quigley2009ros}. ROS and Gazebo enable students to take advantage of various community-developed packages while learning robotics and writing algorithms.


\subsubsection{PX4 and QGroundControl} PX4 is an open-source flight control software that supports aerial, ground, and sea vehicles. It provides capabilities like state estimators, software-in-the-loop simulations, sensor support, position, velocity, and rate controllers. 
QGroundControl is a remote flight monitoring and planning software, that communicates with any MavLINK enabled vehicle. Accessibility to QGroundControl inside the OpenUAV allows students to run simulations closer to how they would operate PX4 vehicles in a field experiment.

\subsubsection{Unity Game Engine and ROS-Sharp} Unity is a cross-platform game engine developed primarily for the creation of 2D, 3D, and virtual reality games \cite{unity}. We can render photorealistic scenes in Unity using its High Definition Rendering Pipeline (HDRP), which includes advanced features like the physically-based sky, subsurface scattering, and translucent materials.

This provides a mechanism to test the field experiment code written in the ROS-Gazebo framework in photorealistic outdoor scenes. The cameras are rendered using the Unity scene while still maintaining the physics from Gazebo. We achieve this by replicating all the Gazebo robot models' pose as their corresponding pose in Unity and disabling Unity's physics and collisions for those models.

ROS-Sharp is a ROS communication package in C\# using web sockets \cite{bischoff2017ros}. ROS-Sharp enables us to communicate with the ROS nodes in Gazebo from the Unity environment. ROS-Sharp also provides the necessary conversions from the right-handed coordinate system (Gazebo) to the left-hand coordinate system (Unity). 

\begin{figure}[] 
\centering
\vspace{6pt}
\includegraphics[width=0.48\textwidth]{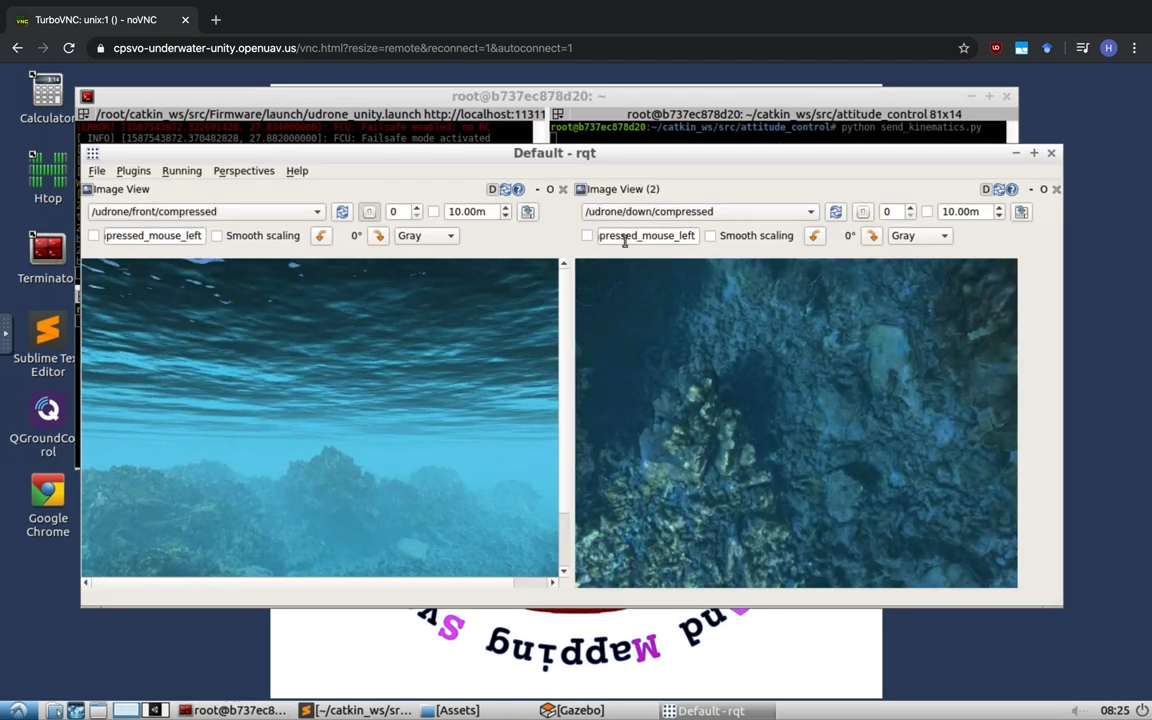}\\
\vspace{0.1 cm}
\includegraphics[width=0.48\textwidth]{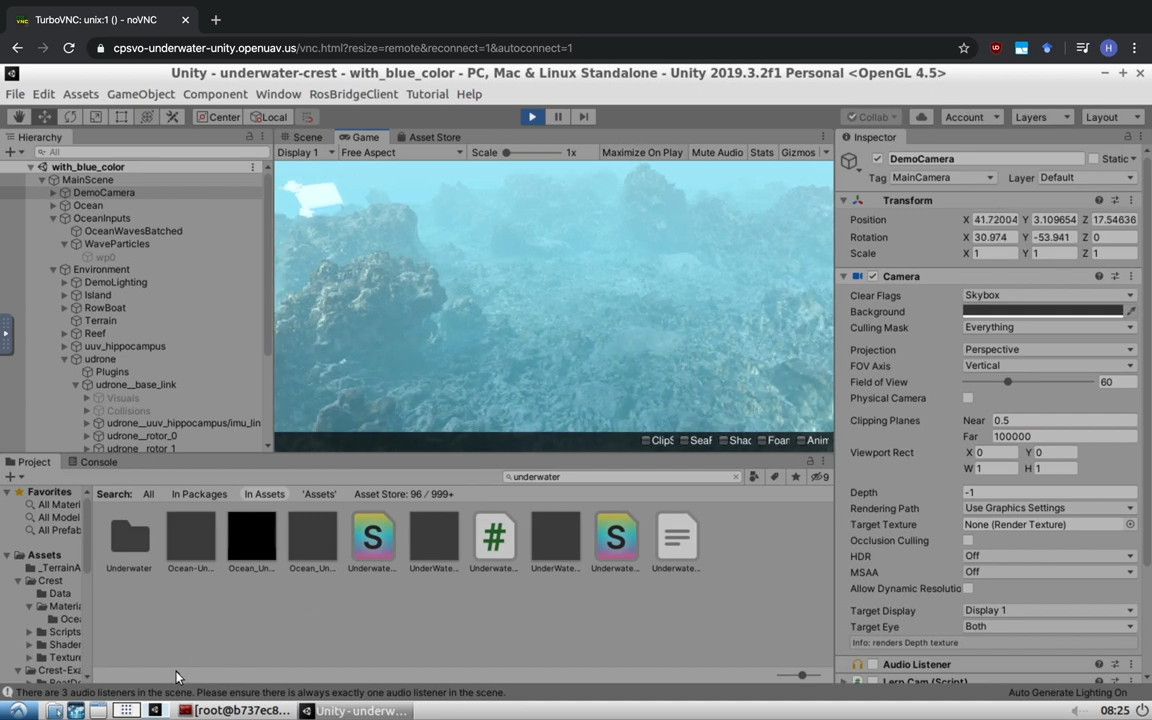}
\caption{Unity rendering of an underwater environment in OpenUAV: The top image shows the rendering of underwater coral reefs from the front and down cameras of an AUV. The bottom image shows Unity updating the kinematics of the AUV game object as per the Gazebo model states.}
\label{fig:unity}
\end{figure}

We use ROS-Sharp to send pose and camera messages between Gazebo and Unity. The ROS-Sharp package can also read URDF and SDF robot models into Unity as game objects. In Fig.~\ref{fig:unity}, we show the front and down camera view from the autonomous underwater vehicle (AUV) using ROS image viewer and the rendering of an underwater coral reef environment in Unity. 

\begin{figure*} 
\centering
\vspace{6pt}
\includegraphics[width=0.75\textwidth]{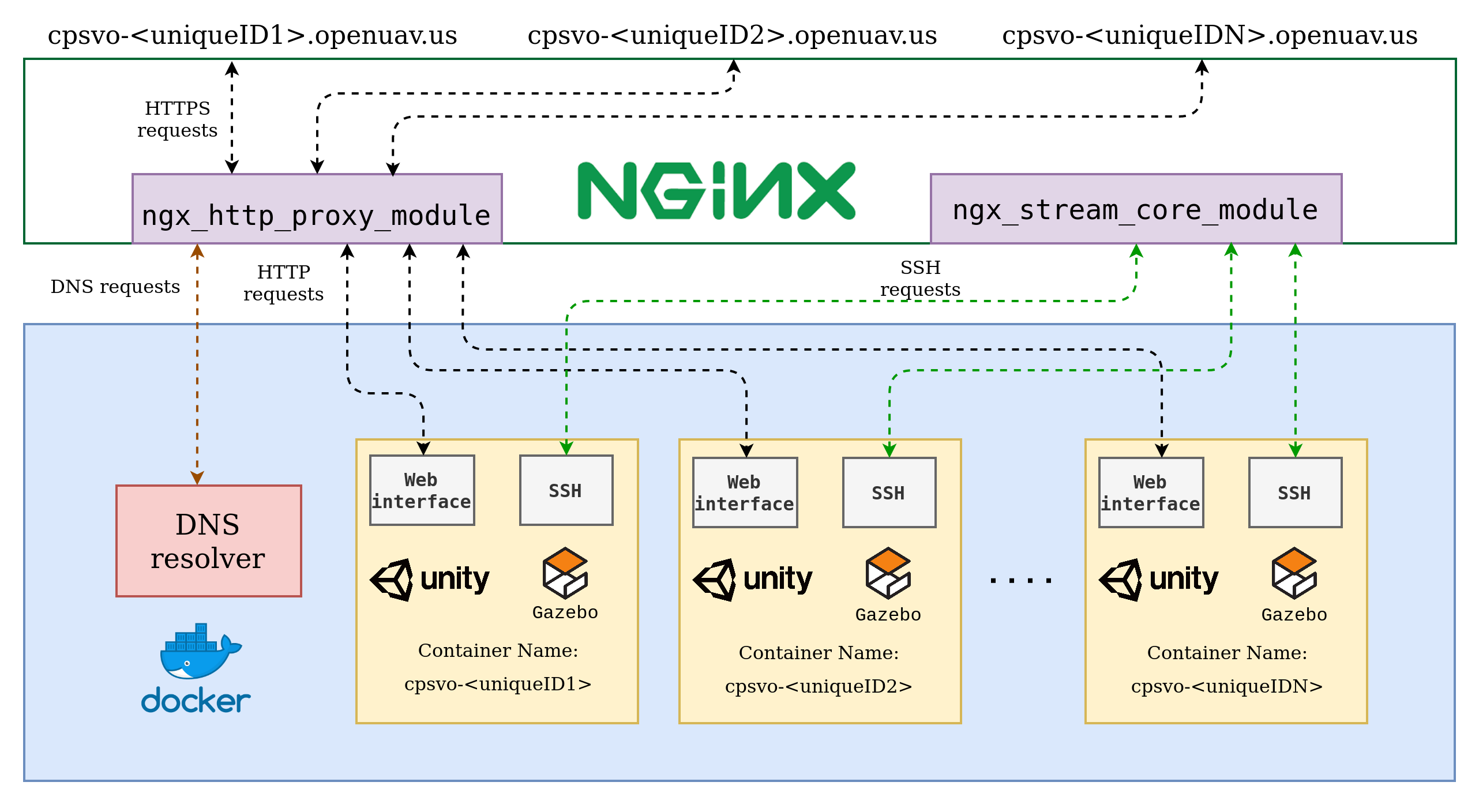}
\vspace{-6pt}
\caption{Nginx HTTP proxy and stream modules are used to proxy multiple HTTP sessions and SSH sessions to outside users, respectively. A Domain Name System (DNS) resolver resolves the URL prefix \url{cpsvo-<uniqueID>} name to its corresponding Docker network IP address. Web interface displays a Lubuntu desktop through NoVNC, and applications like Unity and Gazebo are accessed through this interface.}
\label{fig:openuav-unity} 
\end{figure*}

\subsection{Virtualization Components} 
In OpenUAV, we have operating system level virtualization to deliver software as packages called containers \cite{boettiger2015introduction}.

\subsubsection{Docker} Container technology has become the preferred means of packaging and deploying applications. Docker orchestration tool provides a necessary mechanism to package the software, its libraries, and its dependencies into a single container. 


All OpenUAV containers have a shared network namespace and a Xorg server (X server) with access to GPUs for doing 3D renderings (GLX calls). The X server is a window display server in Linux, which must be available for the containers. We found an X server running on a GPU (Google Cloud Tesla P4) got 558 frames per second (FPS) compared to an X server on CPU with 20 FPS in glxspheres, a graphics performance benchmark tool \cite{glxspheres}.  

\subsubsection{NGINX} Nginx is a web server used widely for hosting files, and load balancing \cite{reese2008nginx}. Nginx does not have a domain name system (DNS) mapping of the container name to its assigned IP address. Nginx is provided with a DNS resolver that can resolve the docker container name to its corresponding IP address. 

User identity management, authentication, and access control are handled by Cyber-Physical Systems Virtual Organization (CPS-VO), similar to how we did in OpenUAV1 \cite{cps, schmittle2018openuav_1}. We expect students and researchers to briefly describe their work when requesting to use our open-source platform. 
After authentication with CPS-VO, each user can create an OpenUAV container. CPS-VO creates a simulation container in the OpenUAV server using Ansible \cite{ansible}. The URL to access this container's desktop is as follows \url{cpsvo-<uniqueID>.openuav.us}, where uniqueID is generated using a Universally Unique IDentifier (UUID) function. The Nginx proxies the URL request to the correct container where the container's name is \url{cpsvo-<uniqueID>}. Nginx's DNS resolver can resolve this container name to its corresponding IP address and serve the remote desktop. The DNS-based approach results in reduced port management on the server machine compared to OpenUAV1.  Fig.~\ref{fig:openuav-unity} shows how each of these components interacts with each other. We expose TCP streaming ports (ngx\_stream\_core\_module in Fig.~\ref{fig:openuav-unity}) for SSH access as per specific user requirements since it takes up a dedicated port on the server machine.

\subsection{Interactive Components} 
Here we describe the software components that enable users to interact with the simulations running in containers. Some of these technologies are used in remote computing and High-Performance Computing (HPC) simulation services \cite{meier2014visualization}.

\subsubsection{VirtualGL and TurboVNC} 

VirtualGL enables the redirection of the OpenGL 3D rendering commands (GLX calls) made by applications like Gazebo or Unity to a GPU, and the rendered 2D framebuffer image gets displayed on a Virtual Network Computing (VNC) virtual display \cite{virtualGL}. 

TurboVNC and VirtualGL provide a high-performing and robust solution for displaying 3D applications over different types of networks \cite{turbovnc, deboosere2007thin}. A study conducted comparing TurboVNC to other VNC options shows that TurboVNC’s JPEG compression algorithm provides faster rendering \cite{TurboVNCspeed}. 
\begin{figure}
\centering
\vspace{6pt}
\includegraphics[width=0.5\textwidth]{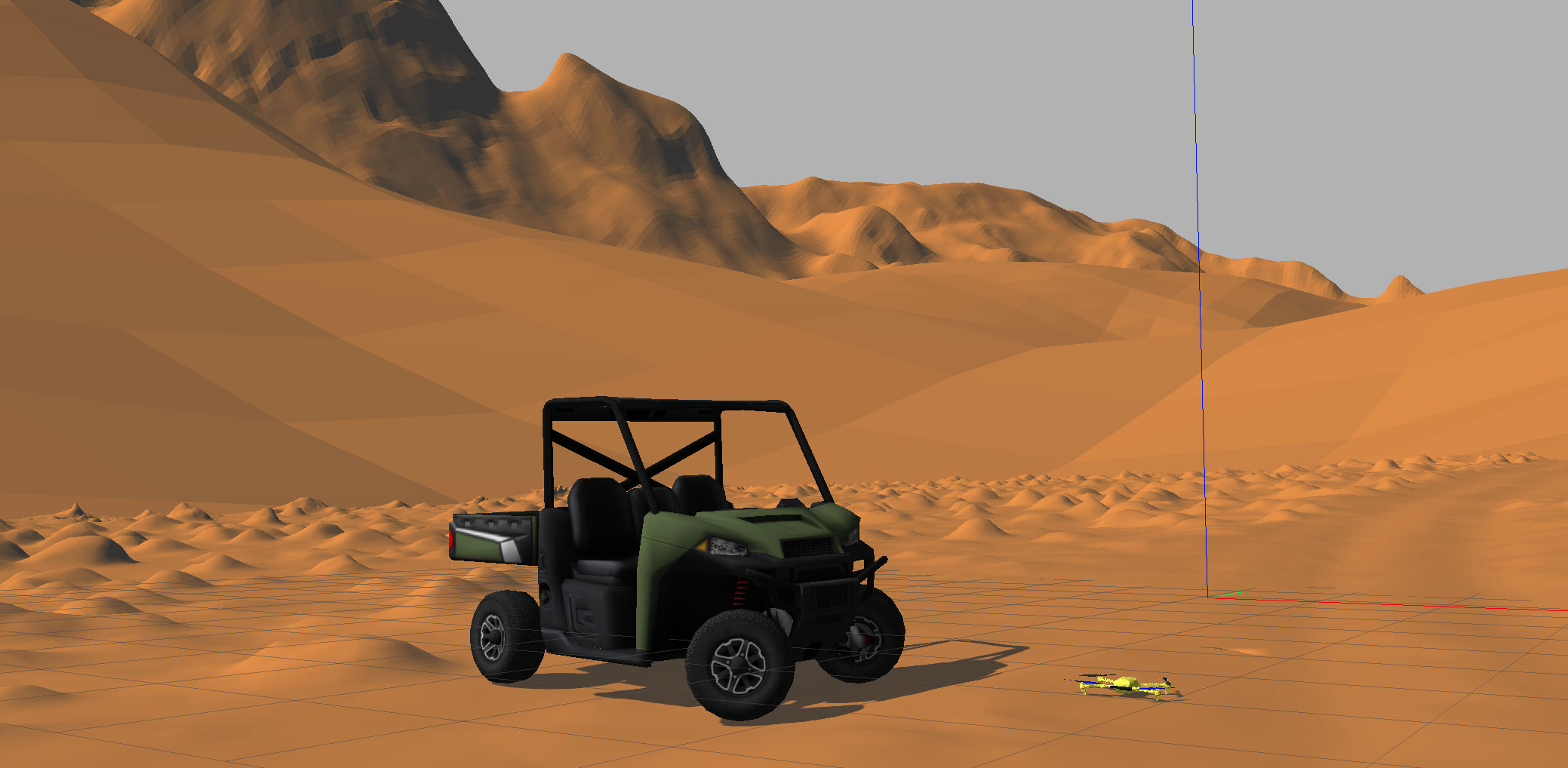}
\caption{\label{fig:environment}{\small}
The simulated environment containing the Mars Jezero world, the rover, and the UAS.}
\label{fig:cps_challenge}
\end{figure}

\begin{figure} 
\centering
\includegraphics[width=0.5\textwidth]{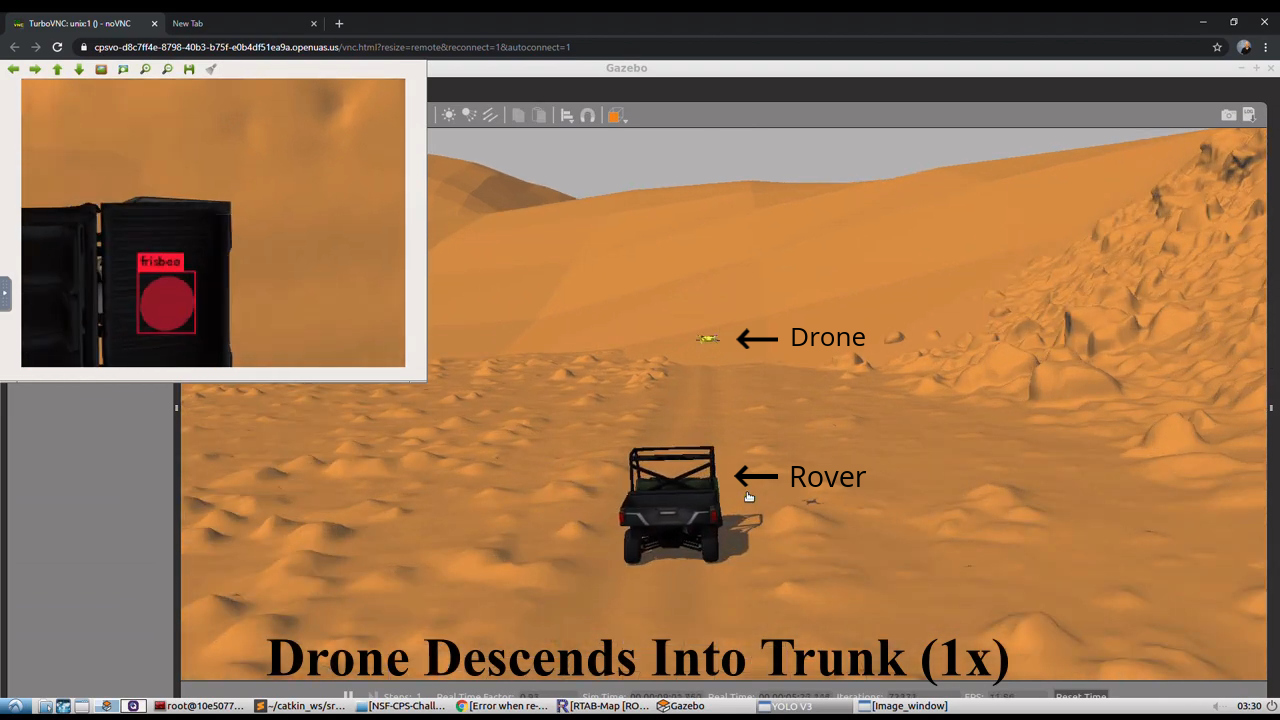}
\caption{Demonstration of phase 2 implementation by a team from ASU during NSF CPS Challenge 2020 in OpenUAV platform \cite{swastik}.}
\label{fig:swastik}
\end{figure}

\begin{figure*}
\centering
\vspace{6pt}
\begin{tikzpicture}
    \begin{axis}[
        ylabel={Containers created},
        height=6cm,
        width=\textwidth,
        date coordinates in=x,
        xmin=2020-01-01,
        xmax=2020-10-06,
        xtick distance=30,
        xticklabel style={
            rotate=45,
            anchor=near xticklabel,
        },
        legend style={at={(0.5,-0.35)},
		anchor=north,legend columns=-1},
        xticklabel=\day.\month.\year,
        scatter/classes={%
		NSF={mark=triangle,blue},%
		SES={mark=square,red},%
		SESE={mark=square,violet},%
		Other={mark=o,draw=black}}
    ]
        \addplot[
	scatter,only marks, %
	scatter src=explicit symbolic]
    table[x=x,y=y,meta=label]{ 
x y label
2020-03-10 9 SES
2020-05-15 10 NSF
2020-03-15 7 SES
2020-05-17 7 NSF
2020-05-18 6 NSF
2020-02-09 5 SES
2020-02-04 5 SES
2020-05-21 5 NSF
2020-05-30 13 NSF
2020-06-01 6 NSF
2020-05-20 5 NSF
2020-04-17 5 SES
2020-04-14 4 SES
2020-01-24 4 SES
2020-05-29 13 NSF
2020-05-23 4 NSF
2020-02-10 4 SES
2020-04-01 4 SES
2020-01-25 4 SES
2020-05-01 3 Other
2020-02-06 3 SES
2020-05-16 3 NSF
2020-01-23 3 SES
2020-03-20 3 SES
2020-05-28 7 NSF
2020-03-22 3 SES
2020-01-20 3 SES
2020-03-27 3 SES
2020-10-21 3 Other
2020-01-16 3 SES
2020-02-07 3 SES
2020-09-10 3 Other
2020-02-12 3 SES
2020-07-12 2 Other
2020-02-29 2 SES
2020-01-29 2 SES
2020-04-07 2 SES
2020-03-25 2 SES
2020-02-05 2 SES
2020-05-19 2 NSF
2020-05-22 2 NSF
2020-01-22 2 SES
2020-04-10 2 SES
2020-04-11 2 SES
2020-06-12 2 NSF
2020-03-31 2 SES
2020-04-28 2 SES
2020-02-11 2 SES
2020-09-17 2 Other
2020-02-20 2 SES
2020-04-02 2 SES
2020-03-19 2 SES
2020-05-24 3 NSF
2020-04-06 2 SES
2020-03-02 2 SES
2020-10-09 1 Other
2020-02-13 1 SES
2020-04-09 1 SES
2020-10-16 2 Other
2020-09-18 2 Other
2020-10-26 2 Other
2020-03-24 1 SES
2020-04-20 1 SES
2020-07-25 1 Other
2020-03-17 1 SES
2020-05-07 1 Other
2020-10-02 1 Other
2020-10-20 1 Other
2020-02-21 1 SES
2020-05-11 1 Other
2020-06-11 1 NSF
2020-06-10 1 NSF
2020-01-28 1 SES
2020-02-01 1 SES
2020-03-03 1 SES
2020-05-04 1 Other
2020-05-31 1 NSF
2020-04-27 1 SES
2020-01-21 1 SES
2020-05-08 1 Other
2020-03-18 1 SES
2020-06-02 7 NSF
2020-04-15 1 SES
2020-08-24 1 Other
2020-04-25 1 SES
2020-04-13 1 SES
2020-08-27 1 Other
2020-10-28 2 Other
2020-04-18 1 SES
2020-08-20 1 Other
2020-09-20 1 Other
2020-04-29 1 SES
2020-03-29 1 SES
2020-03-04 1 SES
2020-04-04 1 SES
2020-01-15 1 SES
2020-01-26 1 SES
2020-03-28 1 SES
2020-05-06 1 Other
2020-04-05 1 SES
2020-09-30 1 Other
2020-01-10 1 Other
2020-09-28 1 Other
2020-01-17 1 SES
2020-05-27 2 NSF
2020-03-13 1 SES
2020-06-03 6 NSF
2020-05-14 5 Other
2020-05-13 2 Other
2020-09-12 2 Other
2020-10-15 2 Other
2020-08-10 1 Other
2020-07-10 1 Other
2020-07-29 1 Other
2020-10-12 1 Other
2020-07-30 1 Other
2020-08-28 1 Other
2020-10-14 1 Other
2020-07-14 1 Other
2020-05-26 1 NSF
2020-09-01 16 SESE
2020-08-27 13 SESE
2020-09-08 4 SESE
2020-08-26 3 SESE
2020-08-28 2 SESE
2020-09-29 2 SESE
2020-09-22 2 SESE
2020-09-16 1 SESE
2020-09-15 1 SESE
2020-09-09 1 SESE
2020-08-25 1 SESE
    };
	    \legend{NSF CPS Challenge 2020 , SES 494/598 , SES 230, General Research Use} 
    \end{axis}
\end{tikzpicture}

\caption{OpenUAV containers created through CPS-VO from January 2020 to September 2020. X-axis represents the time period from January 2020 to September 2020 and y-axis represents the number of containers students and researchers created on each day. Students and researchers created a total of 136, 107, 46 and 56 containers during SES 494/598 course, NSF CPS Challenge 2020, SES 230 course and for general research uses respectively.}
\label{fig:cps-vo-plot}
\end{figure*}
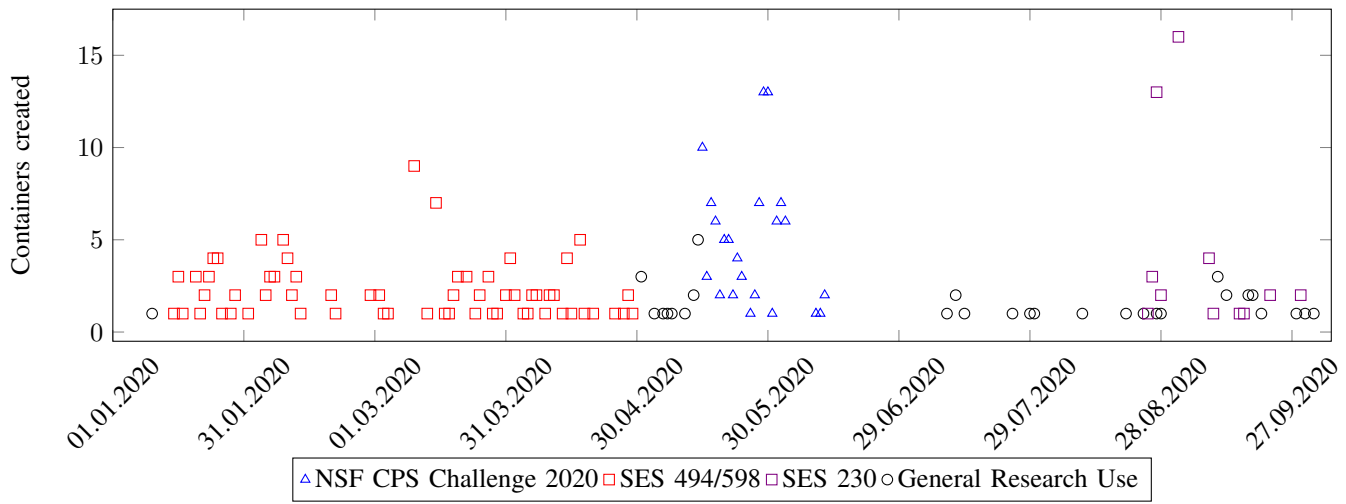

\subsubsection{NoVNC} NoVNC is a JavaScript-based Virtual Network Computing (VNC) application that provides access to VNC sessions over the web browser \cite{martin2015novnc}. NoVNC follows the standard VNC protocol and has support for persistent connection through WebSockets. Nginx proxies the NoVNC client connection to the TurboVNC service of each container.

\subsubsection{Secure Shell FileSystem (SSHFS)} Secure Shell FileSystem allows users to mount a remote file system using the Secure Shell File Transfer Protocol (SFTP) \cite{hoskins2006sshfs}. Through SSHFS, the OpenUAV allows users to utilize their local IDE for code development. This feature requires a port to be exposed on the server and hence is provided based on user requirements.

\section{Use Cases}
\subsection{Cyber-Physical Systems Virtual Organization}
Cyber-Physical Systems Virtual Organization (CPS-VO) is a collaboration among CPS professionals in academia, government, and industry \cite{cps}. OpenUAV was integrated as part of CPS-VO to help students and researchers from various institutions use this robotics platform. CPS-VO provides authentication and authorization for each user's simulations. To maintain a judicious usage of CPU and GPU resources on the OpenUAV server, the CPS-VO lets students suspend a simulation and resume back from the saved simulation. Suspend and resume functionalities in CPS-VO are implemented using Docker's pause and unpause feature. Students and researchers created a total of 345 containers through CPS-VO from January 2020 to September 2020.

\subsubsection{Educational Applications}

A workflow was developed where 35 students taking part in a field robotics course (SES 494/598 Autonomous Exploration Systems) at Arizona State University (ASU) would register and have authorized access to the OpenUAV platform \cite{ses1, openuavcpsvo}. Simulations contained software like Gazebo, ROS, PX4, and QGroundControl configured correctly. We further added support for Python IDE (PyCharm), text editors (Sublime Text 2), and Google Chrome inside every session to improve students' ease of use. 

The SES 494/598 data points in  Fig.~\ref{fig:cps-vo-plot} represent the 136 containers created between January 14, 2020, and April 30, 2020, when the COVID-19 pandemic affected courses. Most of these containers were created for robotics educational purposes during lecture hours and student assignment submissions. We also find more active student engagement during March 8-15, 2020, where course project discussions began  \cite{ses1}.

Through the coursework, we learned that the students needed daily backups to avoid data loss. Utilizing Docker's commit feature, we saved students' work every night as Docker images. Additionally, the CPS-VO workflow was modified to reduce accidental termination of sessions and encouraged the use of the suspend and resume feature. At the end of the coursework, we kept student's work as compressed files, which provided a necessary means to reproduce their work. A survey conducted at the end of the course found 87.5\% of the students were satisfied with the overall experience of OpenUAV. In contrast, others raised concerns regarding web browser (Firefox) inside the container crashing when there were multiple tabs opened.

An introduction to the python programming course (SES 230 Coding for Exploration) also utilized this platform for programming and data science education \cite{cpsc2}. We created a Docker image specific for these students that contained basic python programming environments and deployed it in Google Cloud Platform. 

\subsubsection{NSF CPS Challenge 2020}

NSF CPS Challenge is a yearly outdoor competition held at TIMPA Airfield, Arizona \cite{cpsc1, cpsc2}. This year's competition was made completely virtual, as it happened during the COVID-19 pandemic period. The NSF CPS Challenge 2020 focused on a Mars 2020 theme, and used the Perseverance, and Ingenuity rover and drone duo \cite{mars2020} as shown in Fig.~\ref{fig:cps_challenge}. Teams from universities worldwide would have to simulate an autonomous science probe deployment and recovery mission involving a heterogeneous team of a rover and drone at the Jezero crater landing site. 

During phase 1 of the competition, the teams wrote code for probe deployment from a drone at a particular location, and phase 2 had the teams recover these probes using a drone and land on a moving rover. Fig.~\ref{fig:swastik} shows the phase 2 implementation from one of the competition teams. 


An easy-to-use, competition-specific Docker image was created containing the Jezero crater model, PX4 vehicles, sensor probe, and sample codes. We saved each teams containers as compressed files for the reproducibility of their submissions. The web-based architecture enabled students from universities in India and Argentina to simulate various environments without powerful local compute machine requirements.

Each team score was based on 20 trials of their implementation on deploying the soil probe, maneuvering and retrieving the soil probe, and returning and landing the sUAS on the rover. These 20 trials for each team were scripted and ran on a single container. Our work also opens up the possibility for rigorous Monte Carlo styled software testing of heterogeneous robot simulations because of the simulations' containerized nature. Further details regarding the competition and each teams' implementation code can be found here \cite{cpsc1, mick2021robotics}.


\subsection{Research Applications}

\begin{figure}
\centering
\includegraphics[width=0.48\textwidth]{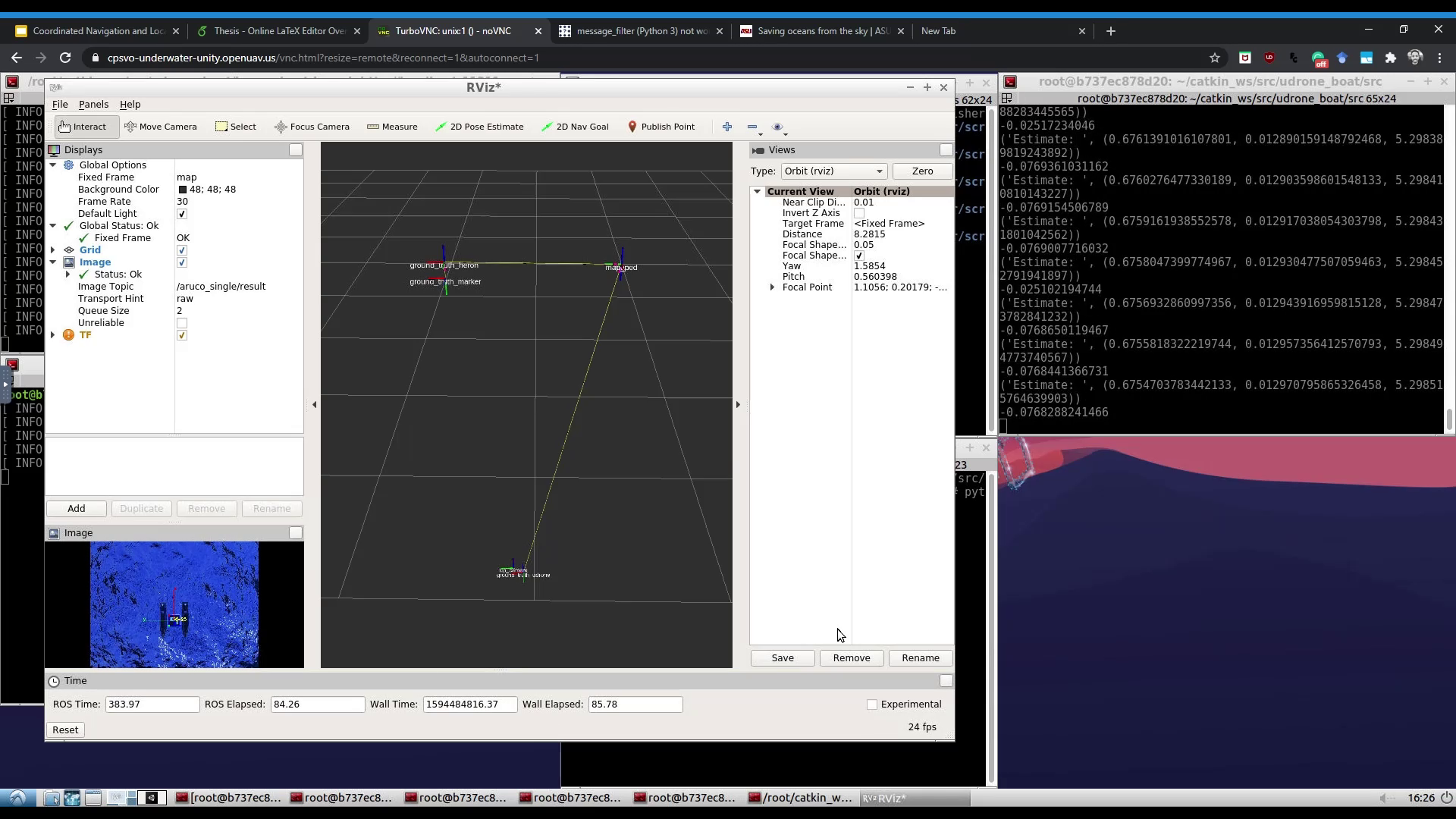}
\caption{An RVIZ window containing the AUV and ASV pose, along with the fiducial marker detections.}
\label{fig:unity1} 
\end{figure}
\subsubsection{Photorealistic underwater coral reef environment}

\begin{figure} 
\centering
\vspace{6pt}
\includegraphics[width=0.45\textwidth]{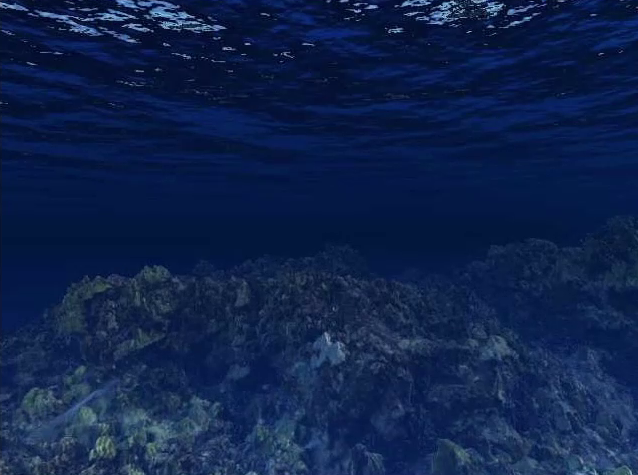}
\caption{The front camera view of the coral reef environment. The coral reef model belongs to Dr. John Burns, Multiscale Environmental Graphical Analysis (MEGA) Lab at the University of Hawai`i, Hilo. }
\label{fig:unity2} 
\end{figure}

A thesis on coordinated navigation and localization of an AUV following an autonomous surface vehicle (ASV) was presented using the OpenUAV framework \cite{hanand}. An aquatic environment was created using the Crest ocean rendering implementation in Unity's HDRP pipeline to address the underwater perception challenges like light attenuation, scattering, reflections, and reef caustics \cite{bowles2017crest}. Fig.~\ref{fig:unity2} shows the implemented underwater environment.

For photorealistic rendering in Crest, the water texture and material characteristics are adjusted to be more realistic by comparing it with actual underwater photos. The coordinated localization module for an AUV is implemented by attaching a fiducial marker of known dimensions to the ASV's underside. The upward-facing Unity camera attached to the AUV game object captures the fiducial marker images. The two major concerns when following this approach are maintaining a metric scale for all Unity game objects and calibrating the Unity cameras to create its camera information as ROS topics. The ground truth AUV and ASV model pose published by Gazebo were accessed to update their corresponding pose in the Unity environment using ROS-Sharp.

The Unity camera images are published through ROS-Sharp, and the controller code used these images to follow the ASV. The Unmanned Underwater Simulation (UUVSim) plugin in Gazebo provided the underwater physics for the AUV and ASV, while the camera rendering was enhanced through Unity support. Fig.~\ref{fig:unity1} shows the photorealistic simulated environment accessed through browsers. Further implementation details on the localization module, state estimator, and coordinated navigation controller for AUV-ASV simulation can be found in the thesis \cite{hanand}.  

Our approach of using physics from Gazebo and Unity perception has limitations. The major limitation we found is the need to write scripts to model the robot's interaction with the environment in Unity. An example includes the missing Kelvin wake pattern that the ASV generates in the Unity environment. Hence the current photorealism work is suitable for outdoor mapping applications and vision-based UAS algorithms that involve minor interactions with the environment.


\subsubsection{Data Science}

TagLab is an interactive AI-assisted image annotation tool for semantic segmentation of coral reef orthomosaics \cite{pavoni2020state}. The AI-assisted annotation predictions require TagLab software to have access to a GPU machine. OpenUAV containers' native support for GPU libraries converted the TagLab software as an online accessible annotation tool. OpenUAV Docker images with TagLab were replicated for multiple users, who used it to annotate different coral reef species. The online accessibility meant that the users can remotely annotate images on their laptops, which became an immediate requirement during the pandemic. These data science research needs showcase the versatile and easy-to-use nature of the OpenUAV platform for robotics and automation community.

\section{Conclusion and Future Work} 

This paper has presented OpenUAV as an easy-to-use, reproducible, and online platform for the Cyber-Physical Systems community. The paper presented the challenges and improvements made to the prior version of OpenUAV. Through OpenUAV, the competed teams in the NSF CPS Challenge designed and implemented controllers for a heterogeneous team of vehicles. We presented various academic and research use-cases of the OpenUAV platform and described a method to do photorealistic simulations inside OpenUAV. In the future, we plan to improve upon the photorealism limitations and make the platform widely accessible to the robotics and automation community.

{\small
\section*{ACKNOWLEDGMENT}
This work was supported in part by the National Science Foundation award CNS-1521617. We would also like to acknowledge the contributions and suggestions made by students and researchers who participated in the SES 494/598 course, SES 230 course, NSF CPS Challenge, and CPS-VO users. We also like to thank Darwin Mick, Student, ASU, Bryant Grady, Graduate Student, ASU School of Geographical Sciences and Urban Planning, Dr. Roberta Martin, Associate Professor, ASU Center for Global Discovery and Conservation Science, and Dr. John Burns, Assistant Professor, the University of Hawai`i at Hilo.}

{\small
\printbibliography}
\end{document}